%% file: main.tex
\documentclass[11pt]{article}

\usepackage[preprint]{acl}
\usepackage{times}
\usepackage{latexsym}
\usepackage[T1]{fontenc}
\usepackage[utf8]{inputenc}
\usepackage{microtype}
\usepackage{inconsolata}
\usepackage{graphicx}
\usepackage{amsmath}
\usepackage{amssymb}

\title{Auditing Future-Memory Use in Point-in-Time Language Agents}

\author{Rongtian Ye\\
  Department of Computer Science, Aalto University \\
  \texttt{rongtian.ye@aalto.fi} \\}

\begin{document}
\maketitle

\begin{abstract}
Point-in-time (PIT) agents can recall post-cutoff facts even when their external evidence respects a historical cutoff. We examine whether this access predicts evidence gathering in 48 historical company-comparison tasks. Independent closed-book revenue probes identify the better-remembered company for each model on tasks with unequal recall, fixing the predicted direction before agent runs. We then compare paired runs with real or anonymized company identities while keeping external evidence PIT-valid. Both Qwen and DeepSeek recalled future facts, but neither supported the prespecified shift toward the better-remembered company. Qwen's four allocation estimates on 28 unequal-recall tasks were negative, with descriptive intervals spanning zero. A prospective DeepSeek replication showed an inverse pattern across all four metrics on 20 such tasks, with intervals below zero. These estimates remained negative after each company, pair, and future-fact omission, and all four post-hoc max-T-adjusted sign-flip p-values were below .05. In a prospectively specified follow-up, explicit future facts produced small estimated shifts away from the supplied-fact company, with intervals spanning zero, while final choices shifted toward it. The mechanism behind the natural inverse pattern remains unresolved. These findings support separate assessments of memory accessibility, search allocation, and final outputs alongside cutoff checks on external evidence.
\end{abstract}

\input{sections/01_introduction}
\input{sections/02_related_work}
\input{sections/03_problem_and_audit}

\input{sections/04_experimental_setup}
\input{sections/05_results}

\input{sections/06_discussion}
\input{sections/07_conclusion}
\input{sections/08_limitations}

\bibliography{references}

\appendix
\input{appendix/01_study_design}

\input{appendix/03_task_construction_prompts_retrieval_draft}
\input{appendix/02_audit_details}

\end{document}

%% file: sections/01_introduction.tex
\section{Introduction}
\label{sec:introduction}

Point-in-time (PIT) agents simulate historical decisions using information available before a specified cutoff. In financial analysis and retrospective decision support, access to later documents can invalidate an evaluation even when the resulting answer appears plausible. Recent work on ex-ante inference and leakage-aware backtesting therefore examines the temporal validity of external evidence and model outputs
\citep{liu2026exante,zhang2026temporalleakage,benhenda2026lookahead}.

The historical cutoff constrains retrieved evidence, while post-cutoff facts can remain accessible from the model's parameters. Prior work has studied future-fact recall, identity masking in financial agents, and knowledge-aware search \citep{liu2026exante,zhu2026knowingdoing,feng2026trust}. We ask whether a model's independently measured access to future facts predicts what it chooses to investigate when all external evidence is PIT-valid.

An agent may use a known future outcome to decide which company to investigate first while citing only pre-cutoff documents. It may also recall the outcome without changing its observable search behavior. Final answers and citation dates alone cannot distinguish these cases. We therefore examine the tool calls and evidence selections made during the investigation to test predictions about how future knowledge relates to the agent's behavior.

We define Temporal Policy Contamination (TPC) as a systematic shift in observable information seeking in the direction predicted by independently measured access to post-cutoff facts. Our audit distinguishes three aspects of temporal integrity. Data integrity concerns when external evidence became available; policy integrity concerns whether future knowledge steers observable actions; output integrity concerns whether final claims and citations respect the historical cutoff.

We apply this audit to 48 historical comparisons between public companies using Qwen~\citep{qwen_3_8} and DeepSeek~\citep{deepseekai2026deepseekv4highlyefficientmilliontoken}. Each task has a Normal version with real company identities and an Anonymized version that replaces identifiers while preserving the underlying records, dates, numerical values, and outcome labels. Independent closed-book probes measure each model's recall of the two companies' future revenue. For tasks with unequal recall, we predict that revealing identities will shift evidence gathering toward the better-remembered company. We fix these predictions before collecting trajectories for each model. Four metrics measure allocation during early search, across all actions and decision batches, and among opened sources.

Both models recalled post-cutoff facts, with higher five-probe accessibility in DeepSeek. Neither supported the predicted positive shift. Qwen's four estimates were negative, with descriptive intervals spanning zero. DeepSeek showed a robust inverse pattern: revealing identities reduced the relative allocation to the better-remembered company on all four metrics. This pattern persisted under dependency checks and post-hoc sensitivity analyses. A prospectively specified follow-up supplied explicit future revenue facts to test whether access to future information could substitute for search. Final answers shifted toward the company whose fact was supplied, while evidence allocation showed small estimated shifts away from that company, with intervals spanning zero. The mechanism behind the inverse pattern remains unresolved.

Our contributions are: (1) A directional audit that uses independent measurements of future-fact recall to specify predictions about search allocation before agent runs.
(2) A paired PIT evaluation across two model families, with controlled identity presentation and batch-aware measures of evidence allocation.
(3) Evidence that the prespecified positive allocation shift is unsupported in both models, together with a robust inverse allocation pattern in DeepSeek.
(4) A semantic re-identification audit of anonymized evidence and a prospectively specified follow-up examining how explicit future facts affect search allocation and final choices.

%% file: sections/02_related_work.tex
\section{Related Work}
\label{sec:related}

\paragraph{Temporal contamination.}
Time-aware language models study changing facts \citep{dhingra-etal-2022-time}, while contamination research develops retrieval-based and behavioral diagnostics \citep{deng-etal-2024-investigating}. Prompted-cutoff tests show that earlier-date instructions do not reliably suppress causally related future knowledge \citep{gao-etal-2025-prompts}. Recent methods address temporal leakage through matched controls and planted contamination \citep{zhang2026temporalleakage} and temporal enforcement \citep{zhang2026tempo}. ExAnte combines cutoff-conditioned inference with memory checks to relate future knowledge to answers \citep{liu2026exante}, while TimeSPEC attributes leaked claims to decisions \citep{zhang2026allleaks}.

\paragraph{Parametric knowledge and tool policy.}
Knowledge-conflict studies examine which source a model follows when parametric memory and supplied evidence disagree \citep{longpre-etal-2021-entity,chen-etal-2022-rich}. Post-training can change how latent or contaminated knowledge is expressed \citep{kocyigit2026posttraining,yang2026beyondreasoning}, and misleading-document studies distinguish local verification from end-to-end adoption \citep{zhu2026deepresearch}. Work on adaptive retrieval examines how a model's knowledge informs its search decisions. Mallen et al. (2023) probe factual knowledge to guide retrieval \citep{mallen-etal-2023-trust}, while FLARE uses predicted text and token confidence to decide when to retrieve \citep{jiang-etal-2023-active}. KbPO \citep{feng2026trust} and AdaSearch \citep{lin2025adasearch} learn knowledge-aware retrieval policies. Other approaches use tool/no-tool rollouts or segment-level credit assignment to learn tool policies \citep{chen2026akbe, kumar2026knowing}.

\paragraph{Historical financial agents.}
FinanceHarness evaluates PIT financial research with auditable citations \citep{xiao2026financeharness}. Look-Ahead-Bench \citep{benhenda2026lookahead} and Profit Mirage \citep{li2025profitmirage} study look-ahead bias and controlled contamination in financial tasks. KTD-Fin masks stock and date identities to evaluate financial-agent behavior \citep{zhu2026knowingdoing}. We use paired tasks with real or anonymized company identities, preserving the dates and numerical content of the historical evidence. Closed-book probes measure access to future revenue facts before each model's trajectories are collected. These measurements fix a model-specific prediction: revealing identities will shift the relative allocation of evidence toward the better-remembered company. We test this prediction against the agent's observed search behavior.

%% file: sections/03_problem_and_audit.tex
\section{Problem and Audit Framework}
\label{sec:problem}

Each task \(i\) specifies a historical cutoff \(t_i\) and an eligible evidence set
\[
E_i^{\le t_i}=\{e:\operatorname{available}(e)\le t_i\}.
\]
The agent is asked to predict which of two companies will have higher future revenue growth using this evidence.\footnote{The SEC implementation uses a conservative date-level rule: filing availability must strictly precede the cutoff date (Appendix~\ref{app:growth-labels-cutoffs}).} Post-cutoff revenue determines the evaluation label and is withheld from the PIT tool environment.

Normal (\(N\)) and Anonymized (\(A\)) are paired views of the same canonical task. Normal presents real company identities; Anonymized deterministically replaces typed identifiers while preserving labels and numerical content. Both conditions provide PIT-valid external evidence. The paired comparison measures differences in behavior associated with identity presentation. A separate audit assesses how often company identities can be recovered from the anonymized text (Appendix \ref{app:semantic-reid}). For Qwen, an additional Identity Cue condition supplies the true company names and tickers alongside the Anonymized corpus, allowing us to examine responses to explicit identity cues while holding the evidence corpus fixed.

Future-memory accessibility is measured independently for each model. For model \(r\), task \(i\), and company \(j\in\{1,2\}\), \(m_{ij}^{(r)}\) is the fraction of closed-book probes that recover the company's post-cutoff revenue within 5\% relative error. We define
\[
\begin{aligned}
M_i^{(r)}&=\frac{m_{i1}^{(r)}+m_{i2}^{(r)}}{2},\\
z_i^{(r)}&=\operatorname{sign}\!\left(m_{i1}^{(r)}-m_{i2}^{(r)}\right).
\end{aligned}
\]
Here \(M_i^{(r)}\) summarizes task-level accessibility, and \(z_i^{(r)}\) identifies the better-remembered company; \(z_i^{(r)}=0\) denotes equal probe scores. These directions are fixed before collecting trajectories for that model. Qwen's original confirmatory directions use two probes per fact, while the prospective DeepSeek replication uses five.

An observable trajectory \(\tau_i\) includes the assistant's decisions, tool calls and responses, opened sources, finalization steps, and submitted answer. A decision batch contains all calls issued in one assistant response, before any result from that batch is observed. We treat these calls as simultaneous for analysis. Allocation metrics are computed from the logged investigation actions and tool responses.

For a fixed model \(r\), let \(\mathcal I=\{i:z_i^{(r)}\ne0\}\) denote the evaluated tasks with unequal memory scores. An allocation statistic \(a_k(\tau)\) is signed toward company 1, so larger values indicate greater relative allocation to that company. We define the memory-aligned Normal--Anonymized contrast as
\[
\Delta_k=
\frac{1}{|\mathcal I|}
\sum_{i\in\mathcal I}
z_i^{(r)}
\left[
a_k\!\left(\tau_i^{N,r}\right)
-
a_k\!\left(\tau_i^{A,r}\right)
\right].
\]
Here \(\tau_i^{N,r}\) and \(\tau_i^{A,r}\) are the paired trajectories under Normal and Anonymized. Multiplication by \(z_i^{(r)}\) orients each task's difference toward its better-remembered company, and the contrast averages these differences with equal task weights. Positive values indicate that revealing identities shifts relative allocation toward the better-remembered company; negative values indicate the opposite shift. We apply this contrast to early allocation (K5), overall action allocation, overall batch allocation, and opened-evidence allocation (Appendix \ref{app:metrics}).

Temporal Policy Contamination (TPC) is a systematic shift in observable information seeking in the direction predicted by independently measured access to post-cutoff facts, while external evidence remains PIT-valid. In our company-comparison tasks, the prespecified hypothesis is \(\Delta_k>0\): revealing identities should increase relative evidence allocation to the better-remembered company.

\begin{figure*}[t]
    \centering
    \includegraphics[width=\textwidth]{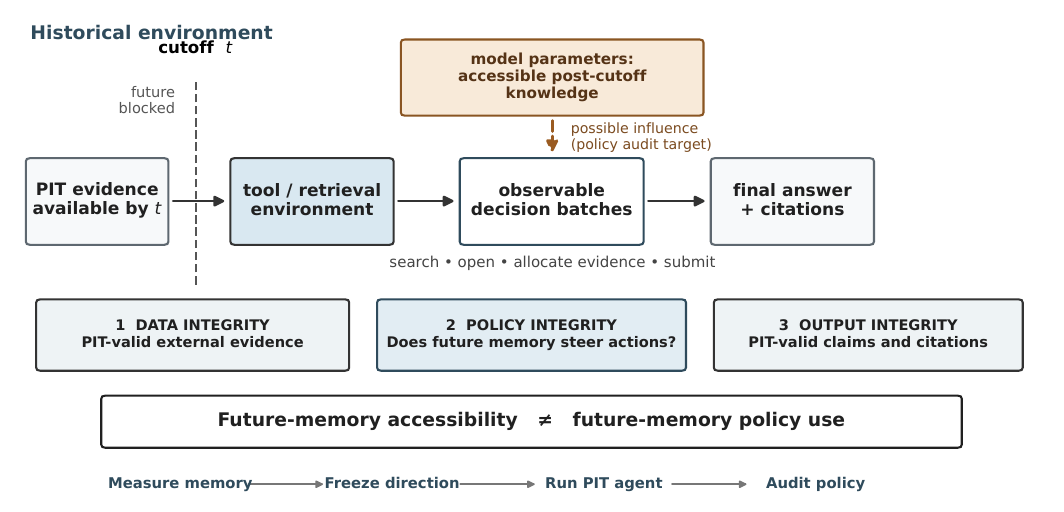}
    \caption{Temporal integrity in PIT agents. The cutoff gate restricts external evidence to information available by time \(t\). Independent memory probes specify a predicted direction of information seeking before the agent runs. The audit compares this prediction with logged actions and distinguishes data, policy, and output integrity.}
    \label{fig:audit-framework}
\end{figure*}

Figure~\ref{fig:audit-framework} distinguishes data integrity, which concerns the temporal validity of external evidence; policy integrity, which concerns the relationship between future knowledge and observable actions; and output integrity, which concerns final claims and citations. Our study enforces the evidence cutoff, records final outputs, and tests the directional allocation hypothesis.

%% file: sections/04_experimental_setup.tex
\section{Experimental Setup}
\label{sec:setup}

\subsection{Tasks and paired environments}

Each task asks which of two public companies will have higher future revenue growth, using evidence available by a historical cutoff. We construct the tasks from public filings in the U.S. Securities and Exchange Commission's (SEC) EDGAR system. Annual revenue observations come from the SEC XBRL Company Facts API and are checked against the corresponding annual statements; filing metadata come from the Submissions API \citep{sec_edgar_api}.

The frozen task pool contains 194 tasks covering 19 companies, 34 unordered company pairs, and target years FY2018--FY2024, with 126 unique future revenue facts. A deterministic stratified rule uses Qwen's original two-probe memory scores and the historical heuristic prior's agreement with the ground-truth winner to select 48 policy tasks covering all 19 companies, 32 pairs, and 77 of these facts. The task set was fixed before policy runs (Appendix~\ref{app:study-design}). Appendix~\ref{app:task-construction-prompts} gives the growth-label and cutoff rules, together with a complete worked task.

The corpus contains 453 historical 10-K/10-Q filings from the SEC EDGAR Archives, divided into 49,854 chunks. Normal and Anonymized are generated from the same canonical task and preserve cutoffs, labels, dates, document structure, and numerical content. Anonymized replaces identifiers through a task-local typed lexicon and uses a separate retrieval index. Audits found no known identifier residuals or numerical mismatches, and retrieval results matched on the checked fields for 288 fixed paired queries. A centralized availability gate enforces the cutoff for all model-visible content, including tool errors. Every assistant action and tool response is logged. Retrieval settings and tool interfaces are specified in Appendix~\ref{app:retrieval-configuration}.

\subsection{Model-specific memory and policy audits}

Both audits were run through the provider Parasail\footnote{https://openrouter.ai/provider/parasail} with temperature set to zero and reasoning and fallback disabled. Investigations allowed up to 18 completed decision batches, subject to a 64-call safety fuse. Finalization allowed at most two evidence-free requests, which were excluded from allocation metrics (Appendix \ref{app:protocol}).

The original audit used \texttt{qwen3.8-27b}\footnote{https://openrouter.ai/qwen/qwen3.8-27b}. Each of the 126 future facts was tested with two fixed closed-book paraphrases, yielding 252 requests. A unit-aware scorer accepted answers within 5\% relative error. We froze the resulting memory directions before running 48 Normal and 48 Anonymized trajectories. The confirmatory analysis uses these two-probe directions. A post-hoc five-probe expansion assesses measurement robustness using the original policy trajectories (Appendix \ref{app:qwen-five-probe}).

For the prospective replication, we used \texttt{deepseek-v4-flash-0731}\footnote{https://openrouter.ai/deepseek/deepseek-v4-flash-0731}. Five globally fixed closed-book probes per fact yielded 630 requests. The model-specific memory directions and 20-task asymmetric subset were frozen before any DeepSeek policy trajectory. We stopped the initial policy-v1 run after repeated failures to satisfy the text-only final-answer format. We then froze a structured adapter for post-investigation A/B submission that preserved the existing conversation. The full 96-run schedule was executed afresh under policy-v2, yielding 48 valid Normal and 48 valid Anonymized trajectories. Appendix \ref{app:deepseek-runtime} documents the runtime provenance. The probe templates and investigation and finalization messages appear in Appendix~\ref{app:probe-agent-prompts}.

\subsection{Measurements, diagnostics, and uncertainty}

We label calls and batches by the companies they target: company 1, company 2, both, or neither. K5 measures allocation over the first five available entity-directed batches. Overall action and batch asymmetry summarize allocation across the full investigation at the call and batch levels, respectively; opened-evidence asymmetry measures the sources the agent inspects. Memory-aligned metrics use each model's frozen direction to orient allocation toward the better-remembered company. Secondary outcomes describe investigation length, tool use, parallelism, bilateralism, finalization, and accuracy. Appendix~\ref{app:metrics} defines the complete metric family, including P1 and K3.

A post-hoc synthetic sensitivity analysis retains the Anonymized action sequences from Qwen's 28 memory-asymmetric tasks and varies evaluator-side direction labels across homogeneous and 30\%/50\% sparse grids (Appendix~\ref{app:synthetic-envelope}).

Qwen familiarity probes measure general identity knowledge and historical financial knowledge (Appendix~\ref{app:familiarity}). A separate Identity Cue study adds true company names and tickers to the Anonymized evidence (Appendix~\ref{app:identity-results}). Across 144 scheduled Normal, Anonymized, and Identity Cue trajectories, 143 produced valid decisions; one Normal finalization failure left 47 complete triplets. A post-hoc semantic re-identification audit tests whether models can recover company identities from three frozen Anonymized chunks per entity using standalone open- and closed-set probes (Appendix~\ref{app:semantic-reid}).

A prospectively specified follow-up tests whether explicit future facts alter evidence allocation. Each of the 48 tasks is run in two otherwise identical Anonymized conditions: one supplies Entity A's true target-year revenue, and the other supplies Entity B's. The prompt thus contains a post-cutoff fact, while retrieved evidence remains subject to the historical cutoff. The primary contrast is positive when allocation shifts away from the company whose fact is supplied. Appendix~\ref{app:explicit-fact} gives the design and results.

Comparisons are paired by task. Broad Normal--Anonymized outcomes use 48 task pairs per model; memory-aligned contrasts use 28 asymmetric Qwen pairs and 20 asymmetric DeepSeek pairs. For the main contrasts, we compute 95\% descriptive task-level bootstrap intervals from 10,000 resamples, keeping conditions paired within each resampled task. Tasks share companies, company pairs, and future facts; these intervals summarize resampling variability within the evaluated task set.

%% file: sections/05_results.tex
\section{Results}
\label{sec:results}
\begin{figure*}[t]
    \centering
    \includegraphics[width=\textwidth]{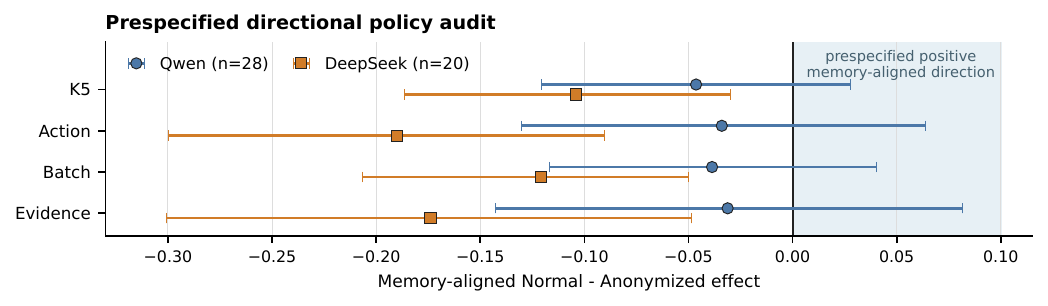}
    \caption{Memory-aligned Normal--Anonymized allocation contrasts. Points show mean paired effects; bars show 95\% descriptive task-bootstrap intervals. Positive values indicate the prespecified shift toward the better-remembered company. Qwen uses its original two-probe directions (n = 28); DeepSeek uses its prospectively frozen five-probe directions (n = 20).}
    \label{fig:knowing-using}
\end{figure*}

\subsection{Memory accessibility and primary allocation results}

Qwen recovered future revenue within 5\% relative error in 124 of 252 original closed-book probes (49.2\%). DeepSeek's prospective five-probe measurement yielded mean accessibility of 0.8413 over the same 126 facts, based on 630 probes.

Figure~\ref{fig:knowing-using} shows the primary memory-aligned Normal--Anonymized contrasts. For Qwen's 28 asymmetric tasks, all four point estimates are negative and all 95\% descriptive intervals span zero: K5 $-0.0463$
$[-0.1207,+0.0276]$, action $-0.0340$ $[-0.1303,+0.0636]$, batch
$-0.0387$ $[-0.1166,+0.0402]$, and opened evidence $-0.0312$
$[-0.1426,+0.0815]$.

DeepSeek shows an inverse allocation pattern across its 20 asymmetric tasks. The corresponding estimates are K5 $-0.1042$ $[-0.1863,-0.0300]$, action $-0.1900$
$[-0.2997,-0.0902]$, batch $-0.1210$ $[-0.2067,-0.0501]$, and opened evidence
$-0.1739$ $[-0.3006,-0.0486]$. All four intervals lie below zero: relative to Anonymized, Normal allocates a smaller share of evidence to the better-remembered company. The prespecified positive shift is unsupported in both models.

\input{tables/table2_diagnostic_evidence.tex}

\subsection{DeepSeek's inverse allocation pattern is robust}

DeepSeek's four allocation estimates remain negative after every prespecified omission of a company, company pair, or future fact. In a post-hoc exact sign-flip analysis under task-level sign symmetry, max-T-adjusted p-values range from .0036 to .0385, with a global max-statistic p-value of .0036 (Appendix~\ref{app:inverse-robustness}).

Continuous signed-memory associations are also negative across the full task set, with Pearson correlations ranging from $-0.283$ to $-0.169$ and Spearman correlations from $-0.377$ to $-0.280$. These analyses use 48 tasks for K5, action, and batch, and 46 for opened evidence. The 28 memory-tied tasks differ from the 20 asymmetric tasks in year and company composition, and 24 ties have both memory scores at 1.0. Directional memory variation therefore remains concentrated in the asymmetric subset.

An exploratory decomposition of absolute allocation also shows greater Normal--Anonymized withdrawal on the better-remembered side. The contrasts are $-0.750$ for early allocation, $-4.650$ for calls, $-1.700$ for batches, and $-3.250$ for opened sources, with all descriptive intervals below zero (Appendix~\ref{app:inverse-robustness}). Qwen does not show the same consistent absolute pattern.

\subsection{Identity cues and output behavior}

Both models use fewer investigation batches and tool calls in Normal than in Anonymized. In the original Qwen audit, mean batches are 13.625 versus 15.750 and mean calls are 26.833 versus 31.313, respectively; accuracy is 36/48 versus 31/48. DeepSeek averages 10.146 versus 13.104 batches and 22.604 versus 28.896 calls, with accuracy of 29/48 in both conditions.

On the asymmetric subsets, the better-remembered company is the ground-truth winner in 15/28 Qwen tasks and 9/20 DeepSeek tasks. Accuracy differences and final-answer alignment show weak, inconsistent associations with the memory measurements (Appendix~\ref{app:output-diagnostics}).

\begin{figure*}[t]
    \centering
    \includegraphics[width=\textwidth]{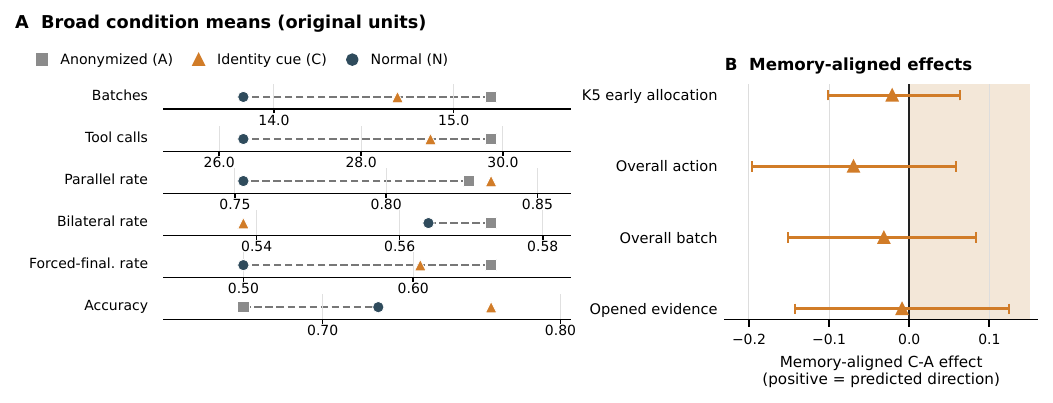}
    \caption{Qwen Identity Cue diagnostic. (A) Mean outcomes for Anonymized (A), Identity Cue (C), and Normal (N). C--A and N--A differences share the same sign on five of six outcomes; parallelism shifts in the opposite direction. (B) Memory-aligned C--A allocation effects with 95\% descriptive task-bootstrap intervals. All four intervals span zero.}
    \label{fig:identity-restoration}
\end{figure*}

In Qwen's separate Identity Cue study, Cue--Anonymized differences have the same sign as Normal--Anonymized differences on five of six broad outcomes; parallelism shifts in the opposite direction (Figure~\ref{fig:identity-restoration}). All four memory-aligned Cue--Anonymized estimates are negative, with descriptive intervals spanning zero.

\subsection{Measurement and mechanism diagnostics}

Qwen's post-hoc five-probe expansion yields mean accessibility of 0.4698 and closely tracks the original memory measurement. Pearson correlations are 0.9077 across 126 facts and 0.9100 across 194 tasks. Of the 28 originally asymmetric policy tasks, 26 retain their direction, one becomes tied, and one changes sign (Appendix~\ref{app:qwen-five-probe}).

Batch allocation is more sensitive than K5 to sparse evaluator-side injections. At $\alpha = .10$, descriptive intervals exclude zero in 99.7\% of homogeneous transformations for K5 and 100.0\% for batch allocation. In the 30\%-sparse setting, these rates fall to 33.5\% and 85.0\%, respectively (Table~\ref{tab:synthetic-envelope}). P1 remains ceiling-limited (Appendix~\ref{app:synthetic-envelope}).

Anonymized context retains substantial information about company identity. In the post-hoc open-set audit, Qwen recovers identities in 63/96 task--entity cases (65.6\%) and DeepSeek in 76/96 (79.2\%). Across DeepSeek's 20 asymmetric tasks, differences in re-identification difficulty between the better- and worse-remembered companies show inconsistent associations with the four memory-aligned allocation effects (Appendix~\ref{app:semantic-reid}).

In the prospectively specified future-fact follow-up, all four allocation estimates point toward withdrawal from the supplied-fact company ($W = 0.0112-0.0386$), with all descriptive intervals spanning zero and a global max-T p-value of .1327 (Appendix~\ref{app:explicit-fact}). Final-answer choices shift toward the supplied-fact company on 17 tasks and away on four, with 27 unchanged. The mechanism underlying the natural inverse allocation pattern remains unresolved.

%% file: tables/table2_diagnostic_evidence.tex
\begin{table*}[t]
\centering
\small
\setlength{\tabcolsep}{3pt}
\caption{Audit stages, analysis status, and principal findings.}
\label{tab:diagnostic-evidence}
\begin{tabular}{p{0.20\textwidth} p{0.25\textwidth} p{0.50\textwidth}}
\hline
\textbf{Analysis} & \textbf{Status} & \textbf{Finding} \\
\hline
Qwen original audit
& Confirmatory; two-probe memory directions
& Four negative N--A estimates ($-0.046$ to $-0.031$); all descriptive intervals span zero. \\
DeepSeek replication
& Prospective confirmatory audit; five-probe memory directions
& Four negative N--A estimates ($-0.190$ to $-0.104$); all descriptive intervals lie below zero. \\
Qwen five-probe expansion
& Post-hoc measurement robustness
& 26/28 original directions retained; all reclassified allocation estimates remain negative. \\
DeepSeek inverse-pattern checks
& Prespecified omissions; post-hoc sign-flip sensitivity
& Every omission preserves the negative sign; max-$T$-adjusted $p$-values range from $.0036$ to $.0385$. \\
Semantic re-identification
& Post-hoc identity audit
& Open-set entity recovery is 65.6\% for Qwen and 79.2\% for DeepSeek; directional associations vary across metrics. \\
Explicit-fact follow-up
& Prospectively specified mechanism follow-up
& Four small positive $W$ estimates ($.011$--$.039$), with intervals spanning zero and global max-$T$ $p = .1327$; final choices shift toward the supplied-fact company. \\
\hline
\end{tabular}
\end{table*}

%% file: sections/06_discussion.tex
\section{Discussion}
\label{sec:discussion}

Both models recovered post-cutoff revenue facts in independent probes, yet neither supported the prespecified shift toward the better-remembered company. Qwen's negative allocation estimates had intervals spanning zero; DeepSeek showed a robust inverse pattern across all four metrics. These findings challenge the hypothesized link between accessible future outcomes and preferential investigation of the associated company.

Search substitution is a plausible explanation for DeepSeek's inverse pattern: access to a future outcome could reduce further investigation of the associated company. The exploratory absolute-count decomposition is consistent with this account, showing greater withdrawal on the better-remembered side (Appendix~\ref{app:inverse-robustness}). A prospectively specified follow-up supplied one company's future revenue explicitly and yielded small shifts in the withdrawal direction, with all intervals spanning zero (Appendix~\ref{app:explicit-fact}). Company identities also remained recoverable from Anonymized text, and relative re-identification difficulty had inconsistent associations with the inverse allocation effects (Appendix~\ref{app:semantic-reid}). The mechanism behind the natural pattern remains unresolved.

Search effort and final decisions responded differently to the experimental conditions. Investigations were shorter in Normal for both models; accuracy was higher for Qwen and unchanged for DeepSeek. In Qwen's Cue study, Cue--Anonymized contrasts shared the sign of Normal--Anonymized contrasts on five of six broad outcomes, while all four memory-aligned allocation intervals spanned zero. The explicit-fact follow-up shifted final choices toward the supplied-fact company while producing weak allocation responses.

These findings motivate PIT evaluations that combine source-availability checks, independent memory probes, trajectory analysis, and assessment of final claims and citations. Behavioral predictions should be fixed before agent runs, with broad investigation measures, memory-aligned allocation, and answer outcomes reported separately. For studies using anonymization as a historical control, semantic re-identification tests complement identifier checks by measuring the identity information retained in the content.

%% file: sections/07_conclusion.tex
\section{Conclusion}
\label{sec:conclusion}

We present a trajectory-based audit that combines independent future-memory probes with paired agent runs under PIT-valid external evidence. Both Qwen and DeepSeek recovered post-cutoff facts, yet neither supported the prespecified shift toward the better-remembered company. Qwen's negative allocation estimates had intervals spanning zero, while DeepSeek showed a robust inverse allocation pattern. PIT evaluation should combine cutoff checks on external evidence with separate assessments of memory accessibility, search allocation, and final outputs.

%% file: sections/08_limitations.tex
\section{Limitations}
\label{sec:limitations}

\paragraph{Scope and dependence.}
Our experiments cover two model families served through Parasail as
\texttt{fp8} endpoints in a 48-task historical-finance setting.
Both models were evaluated on the subset selected using Qwen's
original recall scores and the historical heuristic prior's agreement
with the ground-truth winner.
The effect of deployment precision on memory accessibility remains
unmeasured. The directional analyses use 28 Qwen tasks and 20 DeepSeek
tasks. DeepSeek's 28 tied tasks include 24 with memory scores of
$1.0/1.0$, and their company and year composition differs from the
asymmetric subset. These ties contribute little directional memory
variation to the broader continuous analyses. Repeated companies,
pairs, and future facts induce dependence across tasks. Bootstrap
intervals are descriptive summaries of the evaluated task set;
the post-hoc sign-flip analysis relies on task-level sign symmetry.

\paragraph{Identity and observability.}
The Normal--Anonymized comparison changes identity presentation,
while memory accessibility is measured through independent probes.
The resulting contrasts capture identity-associated behavior, with
future memory as one possible contributor. Company identities remain
recoverable from Anonymized content. The directional re-identification
strata are small, and associations with DeepSeek's inverse pattern
vary across metrics (Appendix~\ref{app:semantic-reid}). Residual identity recovery therefore
remains a potential confound. The logs record investigation actions
and final outputs; the internal use of remembered information remains
unobserved.

\paragraph{Protocol and mechanism.}
DeepSeek's structured finalization adapter was introduced after
policy-v1 format failures and fixed before the complete policy-v2
rerun. The reported DeepSeek analyses use v2 trajectories, with
evidence-free finalization excluded from allocation metrics
(Appendix~\ref{app:deepseek-runtime}). The different finalization protocols limit cross-model
comparisons of stopping behavior. Qwen frequently required forced
finalization, and one Normal-arm finalization failure left 47 complete
triplets in the separate Cue study. P1 is ceiling-limited, and K5 can
miss low-prevalence injections in the synthetic sensitivity analysis
(Appendix~\ref{app:synthetic-envelope}). The explicit-fact follow-up supplies future information
in the prompt and yields small withdrawal estimates with intervals
spanning zero (Appendix~\ref{app:explicit-fact}). The role of parametric future memory in
the natural inverse pattern remains unresolved.

%% file: appendix/01_study_design.tex
\section{Study Design Summary}
\label{app:study-design}

Table~\ref{tab:study-design} lists the designs, samples, and measurements.

\paragraph{Policy-task selection.}
The selector uses Qwen's original two-probe memory scores and
four historical heuristics: latest revenue growth, mean growth over the two
preceding years, growth acceleration, and the majority winner in earlier
available outcomes for the pair. The majority of non-abstaining votes defines
the prior direction, labeled consistent or conflicting by agreement with
the ground-truth winner; balanced or absent votes give a tie.
Prior strength is the absolute A--B vote difference divided by the number
of non-abstaining votes, with bands
weak ($\leq .33$), moderate ($.33$--$.67$, endpoints excluded), strong
($\geq .67$), and missing when all heuristics abstain.

Each task enters the first eligible stratum in this order: S1, $M_i\geq .5$
with a conflicting prior; S2, $M_i\geq .5$ with a consistent prior; S4,
$|m_{i1}-m_{i2}|\geq .5$ and $M_i\geq .25$; S3, $M_i<.5$.
Selection takes twelve tasks per stratum in that order.
At each choice, candidates are ranked lexicographically by new pair,
number of new companies, new target year, number of new future facts,
then lower existing pair, company (maximum then total), and year counts.
For S1/S2, subsequent priorities are both facts scoring 1, both scoring at
least .5, and less represented prior-strength bands; S3 prioritizes less
represented prior relationships. Remaining ties use pseudorandom values
assigned in task-ID order with a fixed seed.

Each candidate must pass a binary feasibility check that the remaining
quotas can still be met: at most two tasks per pair,
eight appearances per company, two uses per future fact, and six positive
and six negative memory differences in S4. This selection was fixed before
corpus construction and policy runs; memory and prior fields remain
evaluator-only. No policy outcome enters the selector.

\renewcommand{\thetable}{A\arabic{table}}
\setcounter{table}{0}
\input{tables/table1_study_design.tex}

%% file: tables/table1_study_design.tex
\begin{table*}[t]
\centering
\small
\setlength{\tabcolsep}{3pt}
\caption{Study designs and samples. N, A, and C denote Normal, Anonymized,
and Identity Cue. Memory-asymmetric samples use the relevant model's fixed
probe directions; Qwen's five-probe reclassification is post-hoc.}
\label{tab:study-design}
\begin{tabular}{p{0.18\textwidth} p{0.27\textwidth} p{0.24\textwidth} p{0.25\textwidth}}
\hline
\textbf{Analysis} & \textbf{Design and conditions} & \textbf{Sample} & \textbf{Measurements} \\
\hline
Qwen memory accessibility & Closed book; two paraphrases per fact, before policy runs & 194 tasks; 126 facts; 252 probes & Recall within 5\%; fact and task memory scores \\
Paired environments & N/A views of one canonical task & 48 tasks; 19 companies; 32 pairs; 77 future facts & Temporal validity; identifier, numeric, and retrieval parity \\
Qwen policy audit & Paired N/A; original two-probe directions & 96 valid trajectories; 28 asymmetric pairs & Memory-aligned K5, action, batch, and evidence contrasts \\
Familiarity diagnostics & Closed-book identity and pre-FY2018 revenue probes & 19 companies; 190 requests & Familiarity scores and familiarity-aligned allocation \\
Synthetic sensitivity & Evaluator-side transformations of fixed A sequences & 28 sequences; 1,000 transformations per grid cell & P1, K5, and batch interval-exclusion rates \\
Identity Cue study & Separate N/A/C runs; A and C share evidence & 144 scheduled; 143 valid; 47 complete triplets & Broad outcomes; raw and memory-aligned C--A allocation \\
Qwen five-probe robustness & Post-hoc measurement expansion; original policy trajectories & 126 facts; 194 tasks; 48 policy tasks & Score/direction agreement; reclassified allocation \\
DeepSeek replication & Prospective five-probe N/A audit; policy-v2 runs & 630 probes; 96 valid trajectories; 20 asymmetric pairs & Memory-aligned allocation and dependency omissions \\
DeepSeek inverse-pattern checks & Post-hoc analysis of existing N/A trajectories & 20 asymmetric tasks; all 48 for continuous associations & Sign-flip sensitivity; task composition; absolute allocation \\
Semantic re-identification & Open-set and candidate-assisted probes for both models & 48 tasks per model; 192 requests across both models and modes & Entity recovery; memory-relative recovery and allocation \\
Explicit future-fact follow-up & Prospectively specified LEAK\_A/LEAK\_B on A environments & 96 valid trajectories; 48 complete pairs & Normalized $W$, absolute $U$, and answer choice \\
\hline
\end{tabular}
\end{table*}

%% file: appendix/03_task_construction_prompts_retrieval_draft.tex
\section{Task Construction, Prompts, and Retrieval}
\label{app:task-construction-prompts}

\subsection{Growth labels and historical cutoffs}
\label{app:growth-labels-cutoffs}

\paragraph{Data sources and provenance.}
The audited annual revenue histories span FY2016--FY2024. Revenue records
from the SEC XBRL Company Facts API are linked by company identifier
(CIK) and accession number to Submissions metadata, including historical
continuation files. Each accepted value is checked against an inline XBRL
or rendered consolidated annual-statement row, retaining the actual
fiscal-period dates, USD value and unit, and original XBRL concept.

The agent corpus uses primary filing texts from the SEC EDGAR Archives:
121 Forms 10-K, 2 Forms 10-K/A, and 330 Forms 10-Q. Item 2.02 8-Ks supply
the full-year outcome-availability proxies and are excluded from the
retrieval corpus. Raw JSON and HTML snapshots are immutable; source URLs,
accession numbers, and SHA-256 hashes link the derived records to their
original responses.

\paragraph{Growth labels.}
For company $j$ and target fiscal year $y$, let $R_{j,y}^{(v)}$ denote
annual consolidated revenue in accounting view $v$. We compute
\[
g_{j,y}^{(v)} =
\frac{R_{j,y}^{(v)}-R_{j,y-1}^{(v)}}{R_{j,y-1}^{(v)}},
\]
using positive revenues for adjacent actual fiscal periods. Each ratio
uses the same accounting view for its numerator and denominator.
The as-reported view uses the original current-year 10-K; the restated
view uses the latest authoritative annual-statement representation in
the frozen source snapshot. Company A precedes Company B in lexicographic
ticker order. The label is A when $g_{A,y}>g_{B,y}$ and B otherwise.
We retain a candidate only when the two views yield the same winner and
\[
\min_{v\in\{\mathrm{as\ reported},\mathrm{restated}\}}
\bigl|g_{A,y}^{(v)}-g_{B,y}^{(v)}\bigr| \geq 0.02.
\]
Thus the gap must be at least two percentage points in both views.
The as-reported winner is the stored ground-truth label. Later restatements
are used to check label stability and are excluded from historical
observations unless independently available before the task cutoff.

Let $a_{j,y-1}$ be the availability date of company $j$'s original
10-K for the base year. The common cutoff is
\[
t_i=\max(a_{A,y-1},a_{B,y-1})+1\text{ calendar day}.
\]
For SEC filings, availability is the later of the filing date and the
calendar date of the acceptance timestamp, or the filing date when no
acceptance timestamp is available. The implemented date-level gate admits
only records with availability strictly earlier than $t_i$; a filing
dated on the cutoff day remains unavailable.

For each company, $t_i$ must precede the first full-year outcome
availability proxy and be no later than 69 calendar days after the start
of the target fiscal period. The outcome proxy is the first post-year-end
SEC Item 2.02 8-K available by the original annual filing. No target-year
quarterly report may be available before $t_i$: the check covers reports
whose period end is on or after the target-period start and before its end.
The companies' target fiscal-year ends must differ by at most 31 days.
Candidates with unresolved accounting facts or missing source provenance
are excluded.

These rules define the 194-task natural pool. The later selection of 48
policy tasks also uses Qwen memory scores and the historical prior's
agreement with the ground-truth winner, as specified in
Appendix~\ref{app:study-design}.

\subsection{Worked task: AMD and Amazon, FY2018}
\label{app:worked-revenue-task}

The frozen policy task \texttt{p2a5-nrg-2018-AMD-AMZN} assigns AMD to A
and Amazon to B. Table~\ref{tab:worked-revenue-task}(a) gives their fiscal
periods and base-year filing dates.

\begin{table}[htbp]
\centering
\small
\setlength{\tabcolsep}{3pt}
\caption{AMD--Amazon FY2018 task.}
\begin{tabular}{lrr}
\hline
\multicolumn{3}{l}{\textit{(a) Fiscal periods and filing dates}} \\
\textbf{Field} & \textbf{A: AMD} & \textbf{B: AMZN} \\
\hline
FY2017 start & 2017-01-01 & 2017-01-01 \\
FY2017 end & 2017-12-30 & 2017-12-31 \\
FY2018 start & 2017-12-31 & 2018-01-01 \\
FY2018 end & 2018-12-29 & 2018-12-31 \\
FY2017 10-K available & 2018-02-27 & 2018-02-02 \\
\hline
\multicolumn{3}{l}{\textit{(b) Revenue (USD billions) and growth}} \\
\textbf{Quantity} & \textbf{A: AMD} & \textbf{B: AMZN} \\
\hline
FY2017, as reported & 5.329 & 177.866 \\
FY2018, as reported & 6.475 & 232.887 \\
As-reported growth & 21.5050\% & 30.9340\% \\
FY2017, restated & 5.253 & 177.866 \\
FY2018, restated & 6.475 & 232.887 \\
Restated growth & 23.2629\% & 30.9340\% \\
\hline
\end{tabular}
\label{tab:worked-revenue-task}
\end{table}

The later base-year filing sets the cutoff to 2018-02-28. Neither company
has a visible target-year quarterly report at that date, and the target
fiscal-year ends differ by two days. The Normal user message is:

\begin{quote}
\small\raggedright
Historical cutoff: 2018-02-28. Company A is ADVANCED MICRO DEVICES INC
(AMD); its latest complete annual period is FY2017. Company B is AMAZON
COM INC (AMZN); its latest complete annual period is FY2017. Using only
the provided historical tools, predict which company will have higher
annual Revenue growth in the next fiscal year. Submit A or B with
citations to opened chunks covering both companies.
\end{quote}

The Anonymized message replaces the two name-and-ticker strings with
\texttt{Company\_A} and \texttt{Company\_B}, respectively; all remaining
wording is identical. The paired corpus applies the corresponding
task-local identity substitutions while preserving dates, numerical
evidence, and document structure.

\paragraph{Evaluator's label calculation.}
Table~\ref{tab:worked-revenue-task}(b) gives the revenues and growth rates
used to determine the stored answer. Target-year revenues, future
disclosure dates, and the label are evaluator-only; they are absent from
the policy prompt and historical tools.

Both views select B, and the smaller growth gap is 7.6711 percentage
points. The recorded first full-year outcome proxies are 2019-01-29 for
AMD and 2019-01-31 for Amazon, both after the cutoff. The as-reported
revenues come from AMD's FY2017 and FY2018 10-Ks, available on 2018-02-27
and 2019-02-08, and Amazon's corresponding filings, available on
2018-02-02 and 2019-02-01. Restated values serve only the stability check;
the historical metric tool returns the base-year as-reported values.

\subsection{Memory probes and agent prompts}
\label{app:probe-agent-prompts}

\paragraph{Closed-book revenue probes.}
Each probe asks for one company's target-year revenue without tools or
retrieval. The original Qwen audit uses P1--P2. The post-hoc Qwen
measurement check adds P3--P5, while the prospective DeepSeek audit uses
all five. Each prompt consists of one opening below followed by the
shared suffix, separated by a blank line. The placeholders are replaced
with the canonical company name and target fiscal year. For the worked
task, these are \texttt{ADVANCED MICRO DEVICES INC} and \texttt{2018},
or \texttt{AMAZON COM INC} and \texttt{2018}.

\noindent\textit{P1}
\begin{quote}
\small\raggedright
This is a historical factual-recall test, not a forecast.

What was \{company\_name\}'s reported annual revenue for fiscal year
\{fiscal\_year\}?
\end{quote}

\noindent\textit{P2}
\begin{quote}
\small\raggedright
Historical fact recall only; do not forecast or estimate.

State \{company\_name\}'s reported full-year revenue in fiscal year
\{fiscal\_year\}.
\end{quote}

\noindent\textit{P3}
\begin{quote}
\small\raggedright
Recall one reported historical accounting fact without forecasting.

For fiscal year \{fiscal\_year\}, what annual revenue did
\{company\_name\} report?
\end{quote}

\noindent\textit{P4}
\begin{quote}
\small\raggedright
Complete this historical factual-recall item from memory only.

\{company\_name\} --- fiscal year \{fiscal\_year\} reported full-year
revenue: \_\_\_\_ billion USD.
\end{quote}

\noindent\textit{P5}
\begin{quote}
\small\raggedright
This question asks for a past reported value, not a prediction.

Report \{company\_name\}'s total full-year revenue for fiscal year
\{fiscal\_year\}.
\end{quote}

\noindent\textit{Shared suffix (unit clarification omitted)}
\begin{quote}
\small\raggedright
Return the answer in billions of US dollars.

If you do not know the historical value from memory, answer UNKNOWN.\\
Do not estimate.

Output exactly one numeric value followed by 'billion USD', or UNKNOWN.
\end{quote}

Scoring compares the parsed, unit-normalized answer with the
as-reported target revenue, with a 5\% relative-error tolerance. For AMD
in the example, the reference value is 6.475 billion USD. The fraction
of successful paraphrases defines the fact's memory score; these scores
and the correct values are withheld from the policy agent.

\paragraph{Investigation system message.}
The following system message accompanies the task message in the paired
policy audits:

\begin{quote}
\small\raggedright
You are a historical financial-research agent. Use only the provided
point-in-time historical environment and its tools. One assistant response
is one decision batch, and multiple independent tool calls in one batch
are allowed. Investigate Company A and Company B as needed, then call
submit\_answer exactly once with A or B and citations to opened evidence
for both companies. The investigation horizon is 18 completed investigation
decision batches; a separate 64-call safety fuse prevents runaway tool
use. At the horizon you will receive an evidence-free finalization
instruction. Do not browse externally, claim unavailable tools, or reveal
private reasoning.
\end{quote}

\paragraph{Finalization messages.}
Qwen receives the following message when investigation reaches its limit:

\begin{quote}
\small\raggedright
The investigation phase is complete. No additional evidence or
investigation tools are available. Based only on the evidence already
observed, provide your final choice. Either call submit\_answer using
already opened citations, or output exactly `FINAL: A` or `FINAL: B`.
\end{quote}

Its format-repair message is:

\begin{quote}
\small\raggedright
Submit the final answer using the required format. No additional evidence
or investigation tools are available. Either call submit\_answer using
already opened citations, or output exactly `FINAL: A` or `FINAL: B`.
\end{quote}

DeepSeek policy-v2 retains the investigation messages and uses a
structured finalization adapter. Its limit-triggered message is:

\begin{quote}
\small\raggedright
The investigation phase is complete. No additional evidence or
investigation tools are available. Submit exactly one final binary choice
through the required submit\_answer tool.
\end{quote}

Its format-repair message is:

\begin{quote}
\small\raggedright
No additional evidence or investigation tools are available. Serialize the
final binary choice through the required submit\_answer tool now.
\end{quote}

The adapter exposes only \texttt{submit\_answer} with one required
\texttt{choice} field in \{\texttt{A}, \texttt{B}\}, disallows additional
fields, and forces that tool. The investigation-time submission interface
below is unchanged. Finalization limits and handling follow
Appendices~\ref{app:protocol} and~\ref{app:deepseek-runtime}.

\subsection{Retrieval and tool interface}
\label{app:retrieval-configuration}

For each company, corpus selection uses filings whose report-period end
falls in the inclusive interval from 730 days before the base fiscal-year
end through that fiscal-year end. A selected filing must also pass the
strict availability gate and be absent from the task's excluded future
source accessions. The allowed forms are 10-K, 10-K/A, 10-Q, and 10-Q/A.
Extracted text blocks are grouped within sections and split at a maximum
of 2,400 characters per source chunk, with no overlap. Anonymization
preserves these chunk boundaries.

Normal and Anonymized each index their own visible chunks. Both use the
same task-local identity normalizer on queries and indexed text. The
index covers each chunk's section heading and content, tokenized into
case-folded letter-led alphanumeric/underscore terms and numeric terms
with optional decimal fractions. Query terms are deduplicated before
scoring. Table~\ref{tab:retrieval-settings} summarizes the retrieval settings.

\begin{table}[htbp]
\centering
\small
\setlength{\tabcolsep}{4pt}
\caption{Retrieval settings.}
\begin{tabular}{ll}
\hline
\textbf{Setting} & \textbf{Value} \\
\hline
Ranker & BM25 \\
$k_1$; $b$ & 1.2; 0.75 \\
Returned hits & At most 10, positive scores only \\
Score ties & Fixed corpus chunk order \\
Snippet length & At most 220 characters \\
Identity normalization & Enabled in both conditions \\
\hline
\end{tabular}
\label{tab:retrieval-settings}
\end{table}

Each hit includes chunk and document IDs, filing form, section,
availability date, score, and a snippet. Its source is authorized again
before the result becomes visible. The tools available during
investigation are:

\begin{itemize}
\item \texttt{list\_documents()} returns the task's authorized filing
metadata and chunk counts.
\item \texttt{search\_documents(query)} returns ranked chunk hits under
the settings above.
\item \texttt{open\_document(document\_id, offset, limit)} lists chunk
IDs and sections for a filing; the default limit is 20 and the maximum
is 50. \texttt{open\_chunk(chunk\_id)} returns the selected chunk's text.
\item \texttt{get\_financial\_metric(entity\_id, metric, fiscal\_year)}
returns audited, cutoff-legal base-year revenue and its source document
ID. Here \texttt{metric} is \texttt{revenue}; target-year values are
unavailable.
\item \texttt{run\_python(code)} evaluates a deterministic arithmetic
expression.
\item \texttt{submit\_answer(answer, citations)} submits A or B with at
least two distinct chunk citations covering both companies. The prompt
instructs the agent to cite evidence it has opened.
\end{itemize}

Unavailable documents and invalid requests return bounded error codes
without source text. The availability gate also applies to listed
metadata, opened chunks, and the sources of numerical metric responses.

%% file: appendix/02_audit_details.tex
\section{Observable Policy Metric Definitions}
\label{app:metrics}

\paragraph{Units and entity attribution.}
A call is one tool invocation; batches follow Section~\ref{sec:problem}.
Entity A/B corresponds to company 1/2, whereas condition A denotes Anonymized.
Attribution uses case-insensitive company aliases at identifier boundaries
in \texttt{search\_documents} queries, the \texttt{entity\_id} argument for
\texttt{get\_financial\_metric}, and logged entity metadata for
\texttt{open\_document}/\texttt{open\_chunk}. Calls can target A, B, or both;
\texttt{list\_documents} and \texttt{run\_python} are neutral, and
unmatched targets are unresolved. A batch is A-only or B-only when only that
entity occurs among its investigation calls, bilateral when both occur, and
neutral otherwise. Neutral and unresolved calls contribute to investigation
length but not entity exposure.

\paragraph{Allocation.}
For entity counts $x_A,x_B$, normalized asymmetry is
$(x_A-x_B)/(x_A+x_B)$. P1 records the first entity-directed batch as A-only,
B-only, or bilateral; its memory-aligned score is $+1$, $-1$, or $0$ for
better-only, worse-only, or bilateral, respectively. K3 and K5 apply the
normalized ratio to the earliest three or five available entity-directed
batches. Each unilateral batch contributes one exposure to its target; a
bilateral batch contributes one to each entity. A trajectory with fewer
than $K$ such batches uses all available batches.
Overall action asymmetry counts executed entity-directed calls, and overall
batch asymmetry counts entity exposure over all entity-directed batches,
with the same bilateral convention. Opened-evidence counts are unique opened
document IDs plus unique opened chunk IDs per entity: repeated opens of one
ID count once, while a document and a chunk from it remain separate units.
Memory alignment and N--A contrasts follow Section~\ref{sec:problem}.

\paragraph{Missing values and secondary measures.}
A zero exposure sum makes normalized allocation undefined, including when
no entity-attributed evidence was opened. P1 is undefined without an
entity-directed batch, and memory-aligned measures are undefined for memory
ties. Each paired contrast requires valid decisions and defined values in
both conditions; undefined cells are omitted metric-wise without imputation.
Parallel rate divides investigation batches containing more than one call
by all investigation batches with an executed call. Bilateral rate divides
bilateral batches by all entity-directed batches; the early version uses
the available first $K$ entity-directed batches. Empty denominators are
undefined. Other measures count investigation batches and calls, opened
10-K/10-Q evidence, finalization outcomes, and correct answers among valid
decisions.

\section{Interaction and Canonicalization Protocol}
\label{app:protocol}

Batch results arrive before the next decision; call-array position does not
establish an observation order within a batch. The protocol permits
18 completed investigation batches and uses a 64-call safety fuse; a batch
that would exceed the fuse is rejected in full.

A valid \texttt{submit\_answer} during investigation is a voluntary submission.
Ordinary no-tool text or an invalid single submission triggers an
evidence-free format-repair request. Reaching the horizon or fuse instead
triggers an evidence-free final-choice request. At most two finalization
attempts are allowed; failure to produce exact \texttt{FINAL: A} or
\texttt{FINAL: B} text, or a valid submission, is recorded as a finalization
failure. DeepSeek uses the choice adapter in Appendix~\ref{app:deepseek-runtime}.
Finalization exposes no new evidence and contributes neither allocation nor
investigation-length counts. Investigation calls emitted with a submission
remain logged but are excluded from pre-submission metrics: their results
were unavailable when the answer was selected.

Logs are append-only. Transport retries repeat the exact request and
conversation state. Only technical failures permit trajectory reruns,
numbered in successive generations. After checking schedule metadata, protocol,
and generation continuity, analysis selects the sole highest-generation record
without inspecting answers, correctness, or effect direction. Retry-only
execution neither starts unrun entries nor retries completed,
finalization-failed, or protocol-invalid trajectories.

An earlier Qwen configuration with a 24-call limit prematurely terminated
parallel investigations. Its 26 debugging records are retained and excluded
from the reported policy results.

\section{Familiarity Diagnostics}
\label{app:familiarity}

Qwen's generic identity familiarity score averages six closed-book probes
per company: two each for name-to-ticker, ticker-to-name, and industry
recognition. Historical financial familiarity averages two paraphrases for
each of two annual revenue facts no later than FY2017. Across 19 companies,
these comprise 114 identity and 76 historical requests (190 total).
Identity familiarity has mean 0.7193, population SD 0.1627, and range
0.3333--1; historical familiarity has mean 0.4211, SD 0.4296, and range 0--1.

At company level, identity and historical familiarity have Pearson/Spearman
correlations of 0.102/0.112 and 0.362/0.372 with mean future-memory scores.
Across policy tasks, their two-company means have Pearson correlations of
0.164 and 0.111 with $M_i$.
For tasks with a uniquely more-familiar entity, identity-aligned N--A K5
and batch effects are 0.0320 $[-0.0453,0.1056]$ and 0.0253
$[-0.0480,0.0931]$ ($n=27$). Historical-familiarity effects are 0.0108
$[-0.0625,0.0834]$ and $-0.0087$ $[-0.0801,0.0658]$ ($n=39$).
All four 95\% descriptive task-bootstrap intervals span zero.
The more-familiar and better-remembered entities differ on 10 identity and
14 historical-familiarity tasks.

\section{Additional Identity-Cue Results}
\label{app:identity-results}

This separate Qwen study ran all three conditions, Normal (N), Anonymized
(A), and Identity Cue (C). Of 144 scheduled trajectories, 143 yielded valid
decisions; one Normal trajectory failed finalization. C--A contrasts use
48 valid pairs, N--A and C--N use 47, and three-condition distances use
47 complete triplets. Condition means use all valid runs within each arm:
47 for N and 48 each for A and C. These N/A runs are separate from the
original Qwen policy audit.

The C--A differences are $-0.521$ batches, $-0.854$ calls, $+0.0074$
parallel rate, $-0.0346$ bilateral rate, $-0.0417$ forced-finalization
rate, and $+0.1042$ accuracy. Five share the sign of the contemporaneous
N--A contrast; parallelism has the opposite sign. The bilateral N--A
contrast is only $-0.0034$.
On the common 47-triplet sample, the C mean is closer to the N mean than
the A mean is for batches and calls, but farther for parallel and bilateral
rates. Task-level distance measures a different quantity:
$|A_i-N_i|-|C_i-N_i|$, with positive values indicating that C is closer.
Its mean is positive for batches and bilateral rate and negative for calls
and parallel rate.

Raw, entity-A-oriented C--A action and batch asymmetry shifts are
$+0.0933$ and $+0.0885$, with descriptive intervals above zero. In contrast,
all four memory-aligned C--A means on the 28 asymmetric tasks are negative,
with 95\% descriptive task-bootstrap intervals spanning zero.

\section{Qwen Dependency and Finalization Checks}
\label{app:dependency}

Qwen's K5 estimate remains negative after every leave-one-company,
leave-one-pair, and leave-one-future-fact omission, with ranges
$[-0.0707,-0.0229]$, $[-0.0639,-0.0322]$, and $[-0.0639,-0.0264]$.
Batch effects also remain negative. One company omission moves action to
$+0.0010$, and some MSFT omissions move evidence slightly positive.

The K5 estimate is $-0.0463$ for all valid asymmetric pairs ($n=28$),
$-0.0482$ when both conditions reach K5 ($n=21$), $-0.0473$ in pairs with
matched finalization modes ($n=14$), and $-0.0349$ when neither condition is
forced to finalize ($n=5$).
Appendix~\ref{app:synthetic-envelope} reports the synthetic sensitivity checks.

\section{Qwen Five-Probe Memory Robustness}
\label{app:qwen-five-probe}

This post-hoc measurement check reuses the original policy trajectories;
the primary analysis uses the original two-probe directions. Across 126
facts, two- and five-probe scores have Pearson/Spearman correlations
0.9077/0.9061; across 194 task-level means, the correlations are
0.9100/0.9051. Of the original 28 asymmetric policy tasks, 26 retain the
same sign, one becomes tied, and one flips. Ten original ties become
asymmetric, giving 37 tasks with nonzero five-probe directions.
Table~\ref{tab:qwen-five-probe} reorients the unchanged raw N--A contrasts
by these directions. Evidence uses 36 pairs because the Anonymized run of
the FY2018 MSFT--NKE task opened no entity-attributed document or chunk,
leaving its evidence ratio undefined.
Every leave-one-template-out direction either agrees with the full
five-probe sign or becomes tied; none reverses it.

\begin{table}[t]
\centering
\small
\setlength{\tabcolsep}{3pt}
\caption{Post-hoc Qwen five-probe reclassification of unchanged policy
contrasts. $n$ counts defined paired contrasts; intervals are 95\%
descriptive task-bootstrap intervals from 10,000 resamples.}
\label{tab:qwen-five-probe}
\begin{tabular}{lrrr}
\hline
\textbf{Metric} & \textbf{$n$} & \textbf{N--A} & \textbf{95\% interval} \\
\hline
K5 & 37 & $-0.0485$ & $[-0.1172,+0.0222]$ \\
Action & 37 & $-0.0343$ & $[-0.1180,+0.0517]$ \\
Batch & 37 & $-0.0379$ & $[-0.1092,+0.0345]$ \\
Evidence & 36 & $-0.0145$ & $[-0.1138,+0.0827]$ \\
\hline
\end{tabular}
\end{table}

\begin{table*}[!t]
\centering
\small
\setlength{\tabcolsep}{4pt}
\caption{Synthetic sensitivity on 28 fixed sequences. Median is the median
mean allocation difference across 1,000 transformations; Excl. is the
percentage with a 95\% descriptive interval entirely above zero.}
\label{tab:synthetic-envelope}
\begin{tabular}{llrrrr}
\hline
\textbf{Setting} & \textbf{$\alpha$} & \textbf{K5 median} & \textbf{K5 excl.} & \textbf{Batch median} & \textbf{Batch excl.} \\
\hline
Homogeneous & .02 & 0.0119 & 19.7\% & 0.0124 & 79.6\% \\
 & .05 & 0.0311 & 83.5\% & 0.0321 & 100.0\% \\
 & .10 & 0.0638 & 99.7\% & 0.0660 & 100.0\% \\
 & .15 & 0.1008 & 100.0\% & 0.1021 & 100.0\% \\
 & .20 & 0.1368 & 100.0\% & 0.1394 & 100.0\% \\
 & .30 & 0.2156 & 100.0\% & 0.2154 & 100.0\% \\
Sparse 30\% & .10 & 0.0179 & 33.5\% & 0.0182 & 85.0\% \\
 & .20 & 0.0384 & 85.0\% & 0.0390 & 99.5\% \\
 & .30 & 0.0602 & 97.4\% & 0.0614 & 100.0\% \\
Sparse 50\% & .10 & 0.0311 & 77.5\% & 0.0318 & 99.8\% \\
 & .20 & 0.0668 & 99.9\% & 0.0690 & 100.0\% \\
 & .30 & 0.1080 & 100.0\% & 0.1099 & 100.0\% \\
\hline
\end{tabular}
\end{table*}

\section{Output and Accuracy Diagnostics}
\label{app:output-diagnostics}

These post-hoc diagnostics relate each model's fixed memory measurements
to output behavior. Correlations use all 48 original N/A task pairs per
model and the task-level accuracy difference (Normal correct minus
Anonymized correct). For Qwen, Pearson/Spearman correlations are 0.038/0.039
with $M_i$ and $-0.120/-0.084$ with $|m_{i1}-m_{i2}|$. For DeepSeek,
the corresponding correlations are 0.071/$-0.064$ and 0.208/0.154.

Final choices match the better-remembered entity on 15/28 Normal and
14/28 Anonymized Qwen tasks, an N--A difference of $+0.036$; DeepSeek's
counts are 10/20 and 9/20, a difference of $+0.050$.
Separately, the better-remembered entity is the ground-truth winner on
15/28 Qwen and 9/20 DeepSeek tasks.

\section{Synthetic Sensitivity Analysis}
\label{app:synthetic-envelope}

This post-hoc analysis transforms evaluator-side batch labels on the fixed
Anonymized trajectories from Qwen's 28 originally asymmetric tasks.
Independently for each entity-directed batch, probability $\alpha$ replaces
its A-only, B-only, or bilateral label with the better-remembered entity's
unilateral label. Neutral labels, sequence lengths, and batch positions
remain unchanged. The effect is the transformed-minus-original
memory-aligned allocation, averaged over all 28 tasks.

The homogeneous setting applies the transformation to every task.
Each sparse replicate samples 8 of 28 tasks (the rounded 30\% setting) or
14 of 28 (50\%) uniformly without replacement and applies the transformation
only to those tasks. Each cell uses 1,000 seeded transformations and 1,000
task bootstraps per transformation. The bootstrap resamples paired
transformed/original task values and uses the 2.5th and 97.5th percentiles
of their mean differences.

At $\alpha=.10$, P1 is ceiling-limited: its interval-exclusion rates are
29.8\% under homogeneous transformation, 0.6\% at 30\% sparsity, and
3.6\% at 50\% sparsity.

\section{DeepSeek Inverse-Pattern Robustness}
\label{app:deepseek-dependency}
\label{app:inverse-robustness}

\begin{table*}[!t]
\centering
\small
\setlength{\tabcolsep}{4pt}
\caption{DeepSeek inverse-pattern checks on 20 asymmetric tasks. Natural
is the mean memory-aligned N--A contrast. Raw $p$ is two-sided; max-T
adjusts across four metrics. All four metrics stay negative in 16/16
company, 17/17 pair, and 36/36 fact omissions.
$S$ is the mean absolute-count contrast;
its interval is a 95\% descriptive task-bootstrap interval.}
\label{tab:inverse-robustness}
\begin{tabular}{lrrrrr}
\hline
\textbf{Metric} & \textbf{Natural} & \textbf{Raw $p$} & \textbf{Max-T $p$} & \textbf{$S$} & \textbf{$S$ interval} \\
\hline
K5 & $-0.1042$ & .0149 & .0385 & $-0.750$ & $[-1.350,-0.200]$ \\
Action/calls & $-0.1900$ & .0014 & .0036 & $-4.650$ & $[-8.900,-1.350]$ \\
Batch & $-0.1210$ & .0025 & .0170 & $-1.700$ & $[-3.300,-0.450]$ \\
Evidence & $-0.1739$ & .0160 & .0337 & $-3.250$ & $[-6.900,-0.450]$ \\
\hline
\end{tabular}
\end{table*}

\paragraph{Prespecified dependency omissions.}
All four allocation estimates remain negative after each company, pair,
or future-fact omission (Table~\ref{tab:inverse-robustness}).

\paragraph{Post-hoc sign-flip sensitivity.}
For each of DeepSeek's 20 asymmetric tasks, let
$c_{ik}=z_i[a_k(\tau_i^N)-a_k(\tau_i^A)]$ be its contribution to metric $k$.
The statistic is the studentized mean,
$T_k=\bar c_k/(s_k/\sqrt{20})$, with sample SD $s_k$.
The exact reference distribution enumerates all $2^{20}=1{,}048{,}576$
sign vectors, assigning one shared sign to all four metrics within a task
and recomputing the statistics. Raw two-sided $p$ values compare
$|T_k^*|$ with $|T_k|$; max-T values compare
$\max_j|T_j^*|$ with $|T_k|$. The global test compares the reference and
observed maxima, giving $p=.0036$. This reference distribution assumes
task-level sign symmetry; shared companies, pairs, and facts limit its
interpretation beyond the evaluated dependent task set.

\paragraph{Continuous associations and task composition.}
Associations between signed memory difference $m_{i1}-m_{i2}$ and raw
N--A allocation have Pearson/Spearman coefficients of $-.169/-.280$ for
K5, $-.283/-.377$ for action, $-.197/-.345$ for batch ($n=48$ each), and
$-.272/-.322$ for evidence ($n=46$).
The 20 asymmetric and 28 tied tasks differ in memory and composition:
24 ties have scores 1.0/1.0; asymmetric tasks include no FY2021 task and
six FY2024 tasks, whereas tied tasks include eight FY2021 tasks and no
FY2024 task. They cover 16 and 17 companies and 17 and 23 pairs,
respectively. Directional memory variation is concentrated in the
asymmetric subset.

\paragraph{Absolute-count decomposition.}
Let $x_{ikb}^N,x_{ikb}^A$ denote absolute exposure to the better-remembered
entity under N/A, and $x_{ikw}^N,x_{ikw}^A$ exposure to the worse-remembered
entity. The task-level contrast is
\[
S_{ik}=(x_{ikb}^N-x_{ikb}^A)-(x_{ikw}^N-x_{ikw}^A).
\]
Units are first-five-batch exposures, calls, batch exposures, or unique
opened documents plus chunks; bilateral allocations count once per entity.
Table~\ref{tab:inverse-robustness} reports $\bar S_k$ over all 20 asymmetric
tasks, with 95\% descriptive intervals from 10,000 task bootstraps.
Negative $S$ means greater N--A withdrawal on the better-remembered side.
For calls, the N--A changes are $-6.900$ on the better-remembered side and
$-2.250$ on the worse-remembered side ($S=-4.650$).
This decomposition is exploratory; Appendix~\ref{app:explicit-fact} tests
the proposed withdrawal response.

\section{DeepSeek Runtime and Finalization Provenance}
\label{app:deepseek-runtime}

The initial policy-v1 run stopped after its first six trajectories, three
of which failed finalization with \texttt{text\_grammar\_rejected}.
A diagnostic continuation brought the v1 record to 12 terminal trajectories:
five valid and seven rejected. V1 has no derived allocation analysis.

Before the full policy-v2 rerun, a structured final-answer adapter was fixed:
after investigation, it requires an A/B choice through \texttt{submit\_answer},
preserving the conversation without investigation tools or new evidence.
The rerun retained memory directions, the 48 tasks and N/A conditions,
horizon, metrics, bootstrap, and dependency analyses. All 96 entries were
rerun from scratch, yielding 48 valid trajectories per condition and no
finalization, protocol, or technical failures. Reported DeepSeek policy
results use v2.

Finalization is excluded from investigation metrics
(Appendix~\ref{app:protocol}).
Qwen's text/submission finalization and DeepSeek's structured choice adapter
differ, so stopping and voluntary/forced/format-finalization rates are not
compared across models.

\section{Semantic Re-identification Audit}
\label{app:semantic-reid}

\begin{table*}[!t]
\centering
\small
\setlength{\tabcolsep}{7pt}
\caption{Semantic re-identification. Aggregate entity recovery uses 96
task--entity cases per model and mode; ``Open both'' uses 48 tasks.
B--W is the mean recovery-score difference on Qwen's 28 and DeepSeek's 20
asymmetric tasks. The lower panel correlates recovery contrast $H$ with
memory-aligned N--A allocation $c_k$ on DeepSeek's 20 asymmetric tasks;
negative correlation is the predicted confound direction.}
\label{tab:semantic-reid}
\begin{tabular}{lrrrrr}
\hline
\textbf{Model} & \textbf{Open entity} & \textbf{Open both} & \textbf{Closed entity} & \textbf{Open B--W} & \textbf{Closed B--W} \\
\hline
Qwen & 63/96 (.656) & 19/48 (.396) & 88/96 (.917) & +.071 & +.071 \\
DeepSeek & 76/96 (.792) & 32/48 (.667) & 89/96 (.927) & .000 & $-.100$ \\
\hline
\end{tabular}
\vspace{1em}
\setlength{\tabcolsep}{9pt}
\begin{tabular}{llrr}
\hline
\textbf{Mode} & \textbf{Metric} & \textbf{Pearson $\operatorname{corr}(H,c_k)$} & \textbf{Spearman $\operatorname{corr}(H,c_k)$} \\
\hline
Open & K5 & +.050 & +.049 \\
 & Action & $-.140$ & $-.014$ \\
 & Batch & $-.147$ & +.055 \\
 & Evidence & $-.138$ & $-.124$ \\
Closed & K5 & $-.071$ & +.070 \\
 & Action & $-.107$ & $-.203$ \\
 & Batch & +.099 & +.037 \\
 & Evidence & $-.185$ & $-.303$ \\
\hline
\end{tabular}
\end{table*}

The post-hoc audit supplied three fixed, entity-exclusive Anonymized chunks
per entity, checked for explicit identifiers. Open-set prompts allowed
unknown answers; the candidate-assisted closed set supplied possible
identities. One request per task, mode, and model returned both entity guesses:
48 tasks $\times$ two modes $\times$ two models gave 192 requests.
Deterministic scoring normalizes case, Unicode, punctuation, and spacing
before fixed name/ticker alias lookup: $r_{ie}=1$ for a match to the true
identity and $0$ otherwise, including unknown or unmapped answers.

At least one identity was recovered on 44/48 open-set tasks for each model.
Table~\ref{tab:semantic-reid} separates aggregate recovery across all 48
tasks from better-minus-worse (B--W) recovery on the asymmetric subsets.
For DeepSeek's 20 asymmetric tasks, define $H_i=r_{i,w}-r_{i,b}$ and use
$c_{ik}$ from Appendix~\ref{app:inverse-robustness}.
Positive $H_i$ means that the better-remembered entity is harder to recover;
the proposed confound predicts $\operatorname{corr}(H,c_k)<0$.

Open-set $H=-1/0/+1$ strata contain 4/12/4 tasks; closed-set strata contain
1/16/3. Action and evidence have negative Pearson correlations in both
modes, but K5 and batch do not maintain the predicted direction across
modes and correlation measures. These small strata, especially in the
closed set, limit the directional audit; substantial residual identity
recovery may still affect the N--A comparison.

\section{Explicit Future-Fact Mechanism Follow-Up}
\label{app:explicit-fact}

\begin{table*}[!t]
\centering
\small
\setlength{\tabcolsep}{3pt}
\caption{Prospectively specified explicit-future-fact follow-up.
$N_W$ is the number of defined normalized pairs; intervals are 95\%
descriptive task-bootstrap intervals for mean $W$. Absolute $U$ averages
all 48 pairs. Positive $W$ and $U$ indicate withdrawal from the supplied-fact entity.}
\label{tab:explicit-fact}
\begin{tabular}{lrrrrrr}
\hline
\textbf{Metric} & \textbf{$N_W$} & \textbf{Mean $W$} & \textbf{95\% interval ($W$)} & \textbf{One-sided $p$} & \textbf{Max-T $p$} & \textbf{Absolute $U$} \\
\hline
K5 & 43 & +.0202 & $[-.0188,+.0636]$ & .1762 & .3091 & +0.031 \\
Action & 43 & +.0199 & $[-.0294,+.0762]$ & .2410 & .3969 & +0.344 \\
Batch & 43 & +.0386 & $[-.0051,+.0877]$ & .0617 & .1327 & +0.344 \\
Evidence & 25 & +.0112 & $[-.1114,+.1427]$ & .4344 & .6156 & +0.302 \\
\hline
\end{tabular}
\end{table*}

\paragraph{Design.}
Designed after observing the natural inverse pattern, the DeepSeek follow-up
fixed treatments, allocation direction, schedule, and analyses before its
96 trajectories. Each of 48 Anonymized tasks has two runs: LEAK\_A supplies
only Entity A's true target-year revenue in the prompt; LEAK\_B supplies
only Entity B's.
This post-cutoff fact cannot be cited; retrieved evidence remains PIT-gated.
All 96 decisions were valid, with 24 A-first and 24 B-first pairs.

\paragraph{Allocation measures and results.}
Using normalized allocation $a_{ik}$ from Appendix~\ref{app:metrics}, define
\[
W_{ik}=\tfrac12[a_{ik}(\text{LEAK\_B})-a_{ik}(\text{LEAK\_A})].
\]
Positive $W$ indicates movement away from the supplied-fact entity.
Both arm-specific ratios must be defined: K5, action, and batch each use
43 pairs; evidence uses 25. Undefined pairs are omitted metric-wise without
imputation.
Table~\ref{tab:explicit-fact} gives equal-task means and 95\% descriptive
percentile intervals from 10,000 paired-task bootstraps, resampling the
defined task pairs separately for each metric.
All four $W$ estimates are positive, with intervals spanning zero.

The one-sided test uses studentized means of defined $W$ values and
1,000,000 Monte Carlo sign flips, sharing each task's sign across four
metrics with their original missing-value masks fixed. Max-T uses the
maximum signed statistic; the global test compares reference and observed
maxima. Each $p$ is $(b+1)/(1{,}000{,}000+1)$, where $b$ counts reference
statistics at least as large as the observed threshold. Global $p=.1327$.

For absolute exposure $x$, define the task-level withdrawal contrast
\[
\begin{aligned}
U_{ik}=\tfrac12[&x_{ikB}^{\mathrm{LEAK\_A}}+x_{ikA}^{\mathrm{LEAK\_B}}\\
                &-x_{ikA}^{\mathrm{LEAK\_A}}-x_{ikB}^{\mathrm{LEAK\_B}}].
\end{aligned}
\]
This is mean untreated minus mean treated exposure, with the units and
bilateral counting of $S$ (Appendix~\ref{app:inverse-robustness}). Each $U$ mean
uses all 48 pairs, including zero counts. All four means are positive,
with 95\% descriptive intervals from 10,000 paired-task bootstraps spanning zero.

\paragraph{Answer outcomes.}
Accuracy is 36/48 under LEAK\_A and 35/48 under LEAK\_B. The probability
of choosing Entity A is .833 versus .562. Final choices move toward the
supplied-fact entity on 17 tasks, away on four, and remain unchanged on
27.

\paragraph{Relation to the natural pattern.}
Correlations compare explicit withdrawal $W_{ik}$ with natural inverse
allocation $-c_{ik}$.
On naturally asymmetric tasks where both are defined, Pearson/Spearman
coefficients are .076/.253 for action and .056/.231 for batch ($n=17$),
and $-.105/.090$ for evidence ($n=9$). These weak and varying associations,
together with the allocation intervals, leave the natural inverse
mechanism unresolved.

\section{Model and Endpoint Metadata}
\label{app:model-metadata}

Table~\ref{tab:model-endpoints} lists the identifiers in the request/response
records.

\begin{table}[htbp]
\centering
\small
\caption{Model and endpoint identifiers.}
\begin{tabular}{@{}p{.23\columnwidth}p{.71\columnwidth}@{}}
\hline
\textbf{Model} & \textbf{Requested and returned identifier; Parasail endpoint} \\
\hline
Qwen & \texttt{qwen/qwen3.8-27b}\newline
\texttt{qwen/qwen3.8-27b-20260814} \\
DeepSeek & \texttt{deepseek/}\allowbreak\texttt{deepseek-v4-flash-0731}\newline
\texttt{deepseek/}\allowbreak\texttt{deepseek-v4-flash-20260731} \\
\hline
\end{tabular}
\label{tab:model-endpoints}
\end{table}

Both endpoints are recorded as fp8. Requests use temperature 0 with
reasoning, included reasoning, and provider fallback disabled. Memory
probes have no tools; investigation requests use the fixed tool schema
and automatic tool choice, with finalization handled as described in
Appendices~\ref{app:protocol} and~\ref{app:deepseek-runtime}.

%% file: references.bib
@misc{zhu2026knowingdoing,
      title={From Knowing to Doing: A Memory-Controlled Benchmark for LLM Trading Agents on Stock Markets}, 
      author={Taojie Zhu and Wentao Zhao and Rui Sun and Beidi Luan and Jiacheng Lu and Sinuo Wang and Jing Li and Daxin Jiang and Yonghong He and Zuo Bai},
      year={2026},
      eprint={2605.28359},
      archivePrefix={arXiv},
      primaryClass={cs.AI},
      url={https://arxiv.org/abs/2605.28359}, 
}

@misc{sec_edgar_api,
  author = {{U.S. Securities and Exchange Commission}},
  title = {{EDGAR} Application Programming Interfaces ({APIs})},
  year = {2024},
  url = {https://www.sec.gov/search-filings/edgar-application-programming-interfaces}
}

@misc{zhang2026temporalleakage,
      title={Temporal Leakage in LLM Backtesting: Measurement, Validation, and Adjusted Scores}, 
      author={Zeyu Zhang and Bradly C. Stadie},
      year={2026},
      eprint={2608.02985},
      archivePrefix={arXiv},
      primaryClass={cs.LG},
      url={https://arxiv.org/abs/2608.02985}, 
}

@inproceedings{liu2026exante,
    title = "{E}x{A}nte: A Benchmark for Ex-Ante Inference in Large Language Models",
    author = "Liu, Yachuan  and
      Wei, Xiaochun  and
      Shi, Lin  and
      Li, Xinnuo  and
      Zhang, Bohan  and
      Dhillon, Paramveer  and
      Mei, Qiaozhu",
    editor = "Demberg, Vera  and
      Inui, Kentaro  and
      Marquez, Llu{\'i}s",
    booktitle = "Proceedings of the 19th Conference of the {E}uropean Chapter of the {A}ssociation for {C}omputational {L}inguistics (Volume 1: Long Papers)",
    month = mar,
    year = "2026",
    address = "Rabat, Morocco",
    publisher = "Association for Computational Linguistics",
    url = "https://aclanthology.org/2026.eacl-long.72/",
    doi = "10.18653/v1/2026.eacl-long.72",
    pages = "1551--1571",
    ISBN = "979-8-89176-380-7"
}

@misc{zhang2026tempo,
      title={TEMPO: Temporal Enforcement via Mode-Separated Policy Optimization for Trustworthy LLM Backtesting}, 
      author={Zeyu Zhang and Bradly C. Stadie},
      year={2026},
      eprint={2605.18843},
      archivePrefix={arXiv},
      primaryClass={cs.LG},
      url={https://arxiv.org/abs/2605.18843}, 
}

@inproceedings{feng2026trust,
    title = "Trust Within? Seek Beyond? Knowledge Boundary Aware Policy Optimization for Agentic Search",
    author = "Feng, Tao  and
      Jiang, Xinke  and
      Hu, Xinyan  and
      Zhang, Yonggang  and
      Tao, Zhen  and
      Zhang, Wentao  and
      Liu, Boyang  and
      Jiang, Wenhao  and
      Wu, Chao",
    editor = "Liakata, Maria  and
      Moreira, Viviane P.  and
      Zhang, Jiajun  and
      Jurgens, David",
    booktitle = "Proceedings of the 64th Annual Meeting of the {A}ssociation for {C}omputational {L}inguistics (Volume 1: Long Papers)",
    month = jul,
    year = "2026",
    address = "San Diego, California, United States",
    publisher = "Association for Computational Linguistics",
    url = "https://aclanthology.org/2026.acl-long.1276/",
    doi = "10.18653/v1/2026.acl-long.1276",
    pages = "27664--27682",
    ISBN = "979-8-89176-390-6"
}

@misc{chen2026akbe,
      title={Efficient Agentic Reinforcement Learning with On-Policy Intrinsic Knowledge Boundary Enhancement}, 
      author={Dingwei Chen and Zefang Zong and Zhipeng Ma and Leo Luo and Yang Li and Chengming Li and Peng Chen and Jie Jiang},
      year={2026},
      eprint={2605.26952},
      archivePrefix={arXiv},
      primaryClass={cs.CL},
      url={https://arxiv.org/abs/2605.26952}, 
}

@misc{kumar2026knowing,
      title={Knowing When to Ask: Segment-Level Credit Assignment for LLM Tool Use}, 
      author={Abhijit Kumar and Zoey Wu and Mohit Suley},
      year={2026},
      eprint={2605.27788},
      archivePrefix={arXiv},
      primaryClass={cs.LG},
      url={https://arxiv.org/abs/2605.27788}, 
}

@misc{zhu2026deepresearch,
      title={Is Deep Research Reliable? Misleading Knowledge Induces False Conclusions}, 
      author={Pengyu Zhu and Lijun Li and Longju Yang and Sen Su and Jing Shao},
      year={2026},
      eprint={2607.20891},
      archivePrefix={arXiv},
      primaryClass={cs.AI},
      url={https://arxiv.org/abs/2607.20891}, 
}

@misc{li2025profitmirage,
      title={Profit Mirage: Revisiting Information Leakage in LLM-based Financial Agents}, 
      author={Xiangyu Li and Yawen Zeng and Xiaofen Xing and Jin Xu and Xiangmin Xu},
      year={2025},
      eprint={2510.07920},
      archivePrefix={arXiv},
      primaryClass={cs.AI},
      url={https://arxiv.org/abs/2510.07920}, 
}

@misc{kocyigit2026posttraining,
      title={The Impact of Post-training on Data Contamination}, 
      author={Muhammed Yusuf Kocyigit and Caglar Yildirim},
      year={2026},
      eprint={2601.06103},
      archivePrefix={arXiv},
      primaryClass={cs.LG},
      url={https://arxiv.org/abs/2601.06103}, 
}

@misc{yang2026beyondreasoning,
      title={Beyond Reasoning: Reinforcement Learning Unlocks Parametric Knowledge in LLMs}, 
      author={Wanli Yang and Hongyu Zang and Junwei Zhang and Wenjie Shi and Du Su and Jingang Wang and Xueqi Cheng and Fei Sun},
      year={2026},
      eprint={2605.07153},
      archivePrefix={arXiv},
      primaryClass={cs.CL},
      url={https://arxiv.org/abs/2605.07153}, 
}

@misc{lin2025adasearch,
      title={AdaSearch: Balancing Parametric Knowledge and Search in Large Language Models via Reinforcement Learning}, 
      author={Tzu-Han Lin and Wei-Lin Chen and Chen-An Li and Hung-yi Lee and Yun-Nung Chen and Yu Meng},
      year={2025},
      eprint={2512.16883},
      archivePrefix={arXiv},
      primaryClass={cs.CL},
      url={https://arxiv.org/abs/2512.16883}, 
}

@misc{zhang2026allleaks,
      title={All Leaks Count, Some Count More: Interpretable Temporal Contamination Detection and Mitigation in LLM Backtesting}, 
      author={Zeyu Zhang and Ryan Chen and Bradly C. Stadie},
      year={2026},
      eprint={2602.17234},
      archivePrefix={arXiv},
      primaryClass={cs.AI},
      url={https://arxiv.org/abs/2602.17234}, 
}

@misc{xiao2026financeharness,
      title={FinanceHarness: Autonomous Financial Deep Research Framework}, 
      author={Yijia Xiao and Rujun Han and Yanfei Chen and Zifeng Wang and Ke Jiang and Zhongying CuiZhu and Vishy Tirumalashetty and Wei Wang and Burak Gokturk and Tomas Pfister and Chen-Yu Lee},
      year={2026},
      eprint={2607.27853},
      archivePrefix={arXiv},
      primaryClass={cs.CL},
      url={https://arxiv.org/abs/2607.27853}, 
}

@misc{benhenda2026lookahead,
      title={Look-Ahead-Bench: a Standardized Benchmark of Look-ahead Bias in Point-in-Time LLMs for Finance}, 
      author={Mostapha Benhenda},
      year={2026},
      eprint={2601.13770},
      archivePrefix={arXiv},
      primaryClass={cs.AI},
      url={https://arxiv.org/abs/2601.13770}, 
}

@inproceedings{longpre-etal-2021-entity,
    title = "Entity-Based Knowledge Conflicts in Question Answering",
    author = "Longpre, Shayne  and
      Perisetla, Kartik  and
      Chen, Anthony  and
      Ramesh, Nikhil  and
      DuBois, Chris  and
      Singh, Sameer",
    editor = "Moens, Marie-Francine  and
      Huang, Xuanjing  and
      Specia, Lucia  and
      Yih, Scott Wen-tau",
    booktitle = "Proceedings of the 2021 Conference on Empirical Methods in Natural Language Processing",
    month = nov,
    year = "2021",
    address = "Online and Punta Cana, Dominican Republic",
    publisher = "Association for Computational Linguistics",
    url = "https://aclanthology.org/2021.emnlp-main.565/",
    doi = "10.18653/v1/2021.emnlp-main.565",
    pages = "7052--7063"
}

@inproceedings{chen-etal-2022-rich,
    title = "Rich Knowledge Sources Bring Complex Knowledge Conflicts: Recalibrating Models to Reflect Conflicting Evidence",
    author = "Chen, Hung-Ting  and
      Zhang, Michael  and
      Choi, Eunsol",
    editor = "Goldberg, Yoav  and
      Kozareva, Zornitsa  and
      Zhang, Yue",
    booktitle = "Proceedings of the 2022 Conference on Empirical Methods in Natural Language Processing",
    month = dec,
    year = "2022",
    address = "Abu Dhabi, United Arab Emirates",
    publisher = "Association for Computational Linguistics",
    url = "https://aclanthology.org/2022.emnlp-main.146/",
    doi = "10.18653/v1/2022.emnlp-main.146",
    pages = "2292--2307"
}

@article{dhingra-etal-2022-time,
    title = "Time-Aware Language Models as Temporal Knowledge Bases",
    author = "Dhingra, Bhuwan  and
      Cole, Jeremy R.  and
      Eisenschlos, Julian Martin  and
      Gillick, Daniel  and
      Eisenstein, Jacob  and
      Cohen, William W.",
    editor = "Roark, Brian  and
      Nenkova, Ani",
    journal = "Transactions of the Association for Computational Linguistics",
    volume = "10",
    year = "2022",
    address = "Cambridge, MA",
    publisher = "MIT Press",
    url = "https://aclanthology.org/2022.tacl-1.15/",
    doi = "10.1162/tacl_a_00459",
    pages = "257--273"
}

@inproceedings{mallen-etal-2023-trust,
    title = "When Not to Trust Language Models: Investigating Effectiveness of Parametric and Non-Parametric Memories",
    author = "Mallen, Alex  and
      Asai, Akari  and
      Zhong, Victor  and
      Das, Rajarshi  and
      Khashabi, Daniel  and
      Hajishirzi, Hannaneh",
    editor = "Rogers, Anna  and
      Boyd-Graber, Jordan  and
      Okazaki, Naoaki",
    booktitle = "Proceedings of the 61st Annual Meeting of the Association for Computational Linguistics (Volume 1: Long Papers)",
    month = jul,
    year = "2023",
    address = "Toronto, Canada",
    publisher = "Association for Computational Linguistics",
    url = "https://aclanthology.org/2023.acl-long.546/",
    doi = "10.18653/v1/2023.acl-long.546",
    pages = "9802--9822"
}

@inproceedings{jiang-etal-2023-active,
    title = "Active Retrieval Augmented Generation",
    author = "Jiang, Zhengbao  and
      Xu, Frank  and
      Gao, Luyu  and
      Sun, Zhiqing  and
      Liu, Qian  and
      Dwivedi-Yu, Jane  and
      Yang, Yiming  and
      Callan, Jamie  and
      Neubig, Graham",
    editor = "Bouamor, Houda  and
      Pino, Juan  and
      Bali, Kalika",
    booktitle = "Proceedings of the 2023 Conference on Empirical Methods in Natural Language Processing",
    month = dec,
    year = "2023",
    address = "Singapore",
    publisher = "Association for Computational Linguistics",
    url = "https://aclanthology.org/2023.emnlp-main.495/",
    doi = "10.18653/v1/2023.emnlp-main.495",
    pages = "7969--7992"
}

@inproceedings{deng-etal-2024-investigating,
    title = "Investigating Data Contamination in Modern Benchmarks for Large Language Models",
    author = "Deng, Chunyuan  and
      Zhao, Yilun  and
      Tang, Xiangru  and
      Gerstein, Mark  and
      Cohan, Arman",
    editor = "Duh, Kevin  and
      Gomez, Helena  and
      Bethard, Steven",
    booktitle = "Proceedings of the 2024 Conference of the North American Chapter of the Association for Computational Linguistics: Human Language Technologies (Volume 1: Long Papers)",
    month = jun,
    year = "2024",
    address = "Mexico City, Mexico",
    publisher = "Association for Computational Linguistics",
    url = "https://aclanthology.org/2024.naacl-long.482/",
    doi = "10.18653/v1/2024.naacl-long.482",
    pages = "8706--8719"
}

@inproceedings{gao-etal-2025-prompts,
    title = "Can Prompts Rewind Time for {LLM}s? Evaluating the Effectiveness of Prompted Knowledge Cutoffs",
    author = "Gao, Xin  and
      Zhang, Ruiyi  and
      Du, Daniel  and
      Mahindre, Saurabh  and
      Somayajula, Sai Ashish  and
      Xie, Pengtao",
    editor = "Christodoulopoulos, Christos  and
      Chakraborty, Tanmoy  and
      Rose, Carolyn  and
      Peng, Violet",
    booktitle = "Proceedings of the 2025 Conference on Empirical Methods in Natural Language Processing",
    month = nov,
    year = "2025",
    address = "Suzhou, China",
    publisher = "Association for Computational Linguistics",
    url = "https://aclanthology.org/2025.emnlp-main.1049/",
    doi = "10.18653/v1/2025.emnlp-main.1049",
    pages = "20777--20788",
    ISBN = "979-8-89176-332-6"
}

@misc{qwen_3_8,
  author = {{QwenTeam}},
  title = {Qwen 3.8},
  year = {2026},
  url = {https://qwen.ai/blog?id=qwen3.8},
  note = {Accessed: 2026-09-10}
}

@misc{deepseekai2026deepseekv4highlyefficientmilliontoken,
      title={DeepSeek-V4: Towards Highly Efficient Million-Token Context Intelligence}, 
      author={DeepSeek-AI and Anyi Xu and Bangcai Lin and Bing Xue and Bingxuan Wang and Bingzheng Xu and Bochao Wu and Bowei Zhang and Chaofan Lin and Chen Dong and Chenchen Ling and Chengda Lu and Chenggang Zhao and Chengqi Deng and Chengyu Hou and Chenhao Xu and Chenze Shao and Chong Ruan and Conner Sun and Damai Dai and Daya Guo and Dejian Yang and Deli Chen and Donghao Li and Dongjie Ji and Erhang Li and Fang Wei and Fangyun Lin and Fangzhou Yuan and Feiyu Xia and Fucong Dai and Guangbo Hao and Guanting Chen and Guoai Cao and Guolai Meng and Guowei Li and Han Yu and Han Zhang and Hanwei Xu and Hao Li and Haofen Liang and Haoling Zhang and Haoming Luo and Haoran Wei and Haotian Yuan and Haowei Zhang and Haowen Luo and Haoyu Chen and Haozhe Ji and Hengqing Zhang and Honghui Ding and Hongxuan Tang and Huanqi Cao and Huazuo Gao and Hui Qu and Hui Zeng and J Yang and JQ Zhu and Jia Luo and Jia Song and Jia Yu and Jialiang Huang and Jialu Cai and Jian Liang and Jiangting Zhou and Jiasheng Ye and Jiashi Li and Jiaxin Xu and Jiewen Hu and Jieyu Yang and Jin Chen and Jin Yan and Jingchang Chen and Jingli Zhou and Jingting Xiang and Jingyang Yuan and Jingyuan Cheng and Jingzi Zhou and Jinhua Zhu and Jiping Yu and Joseph Sun and Jun Ran and Junguang Jiang and Junjie Qiu and Junlong Li and Junmin Zheng and Junxiao Song and Kai Dong and Kaige Gao and Kang Guan and Kexing Zhou and Kezhao Huang and Kuai Yu and Lean Wang and Lecong Zhang and Lei Wang and Leyi Xia and Li Zhang and Liang Zhao and Lihua Guo and Lingxiao Luo and Linwang Ma and Linyan Zhu and Litong Wang and Liyu Cai and Liyue Zhang and Longhao Chen and MS Di and MY Xu and Max Mei and Miaojun Wang and Mingchuan Zhang and Minghua Zhang and Minghui Tang and Mingming Li and Mingxu Zhou and Minmin Han and Ning Wang and Panpan Huang and Panpan Wang and Peixin Cong and Peiyi Wang and Peng Zhang and Qiancheng Wang and Qihao Zhu and Qingyang Li and Qinyu Chen and Qiushi Du and Qiwei Jiang and Rui Tian and Ruifan Xu and Ruijie Lu and Ruiling Xu and Ruiqi Ge and Ruisong Zhang and Ruizhe Pan and Runji Wang and Runqian Chen and Runqiu Yin and Runxin Xu and Ruomeng Shen and Ruoyu Zhang and Ruyi Chen and SH Liu and Shanghao Lu and Shangmian Sun and Shangyan Zhou and Shanhuang Chen and Shaofei Cai and Shaoheng Nie and Shaoqing Wu and Shaoyuan Chen and Shengding Hu and Shengyu Liu and Shiqiang Hu and Shirong Ma and Shiyu Wang and Shuiping Yu and Shunfeng Zhou and Shuting Pan and Shuying Yu and Songyang Zhou and Tao Ni and Tao Yun and Tian Jin and Tian Pei and Tian Ye and Tianle Lin and Tianran Ji and Tianyi Cui and Tianyuan Yue and Tingting Yu and Tun Wang and W Zhang and WL Xiao and Wangding Zeng and Wei An and Weilin Zhao and Wen Liu and Wenfeng Liang and Wenjie Pang and Wenjing Luo and Wenjing Yao and Wenjun Gao and Wenkai Yang and Wenlve Huang and Wenqing Hou and Wentao Zhang and Wenting Ma and Xi Gao and Xiang He and Xiangwen Wang and Xianzu Wang and Xiao Bi and Xiaodong Liu and Xiaohan Wang and Xiaokang Chen and Xiaokang Zhang and Xiaotao Nie and Xiaowen Sun and Xiaoxiang Wang and Xin Cheng and Xin Liu and Xin Xie and Xingchao Liu and Xingchen Liu and Xingkai Yu and Xingyou Li and Xinyu Yang and Xinyu Zhang and Xu Chen and Xuanyu Wang and Xuecheng Su and Xueyin Chen and Xuheng Lin and Xuwei Fu and YC Yan and YQ Wang and YW Ma and Yanfeng Luo and Yang Zhang and Yanhong Xu and Yanru Ma and Yanwen Huang and Yao Li and Yao Li and Yao Xu and Yao Zhao and Yaofeng Sun and Yaohui Wang and Yi Qian and Yi Shao and Yi Yu and Yichao Zhang and Yifan Ding and Yifan Shi and Yijia Wu and Yiliang Xiong and Yiling Ma and Ying He and Ying Tang and Ying Zhou and Yingjia Luo and Yinmin Zhong and Yishi Piao and Yisong Wang and Yixiang Zhang and Yixiao Chen and Yixuan Tan and Yixuan Wei and Yiyang Ma and Yiyuan Liu and Yonglun Yang and Yongqiang Guo and Yongtong Wu and Yu Wu and YuKun Li and Yuan Cheng and Yuan Ou and Yuanfan Xu and Yuanhao Li and Yuduan Wang and Yuehan Yang and Yuer Xu and Yuhan Wu and Yuhao Meng and Yuheng Zou and Yukun Zha and Yunfan Xiong and Yupeng Chen and Yuping Lin and Yuqian Cao and Yuqian Wang and Yushun Zhang and Yuting Yan and Yutong Lin and Yuxian Gu and Yuxiang Luo and Yuxiang You and Yuxuan Liu and Yuxuan Zhou and Yuyang Zhou and Yuzhen Huang and ZF Wu and Zehao Wang and Zehua Zhao and Zehui Ren and Zekai Zhang and Zhangli Sha and Zhe Fu and Zhe Ju and Zhean Xu and Zhenda Xie and Zhengyan Zhang and Zheren Gao and Zhewen Hao and Zhibin Gou and Zhicheng Ma and Zhigang Yan and Zhihong Shao and Zhixian Huang and Zhixuan Chen and Zhiyu Wu and Zhizhou Ren and Zhongyu Wu and Zhuoshu Li and Zhuping Zhang and Zian Xu and Zihao Wang and Zihua Qu and Zihui Gu and Zijia Zhu and Zilin Li and Zipeng Zhang and Ziwei Xie and Ziyi Gao and Ziyi Wan and Zizheng Pan and Zongqing Yao},
      year={2026},
      eprint={2606.19348},
      archivePrefix={arXiv},
      primaryClass={cs.CL},
      url={https://arxiv.org/abs/2606.19348}, 
}
